\documentclass{article}
\PassOptionsToPackage{square,sort,comma,numbers}{natbib}
\usepackage[preprint]{neurips_2019}

\usepackage[utf8]{inputenc} 
\usepackage[T1]{fontenc}    
\usepackage{booktabs}       
\usepackage{amsfonts}       
\usepackage{nicefrac}       
\usepackage{microtype}      
\usepackage{multirow}
\usepackage{amsmath}

\usepackage{graphicx}
\usepackage{subcaption}

\DeclareMathOperator*{\ynames}{y_{\text{names}}}
\DeclareMathOperator*{\volume}{\text{vol}(i)}
\DeclareMathOperator*{\ticketmax}{B_{\text{ticket}}}
\DeclareMathOperator*{\namemax}{B_{\text{name}}}
\DeclareMathOperator*{\sectormax}{B_{\text{sector}}}
\DeclareMathOperator*{\irange}{\forall i = 1,\dots,n}

\newcommand*{\ZZ}{\mathbb{Z}}

\newcommand*{\integral}{\mathcal{I}}
\newcommand*{\predc}{\hat{c}}
\newcommand*{\truec}{c}
\newcommand*{\predcmodel}{f_\theta}
\newcommand*{\features}{\phi}
\newcommand*{\optsol}{x^*}
\newcommand*{\MIPLayer}{MIPaaL}

\title{MIPaaL: Mixed Integer Program as a Layer}

\author{
  Aaron Ferber\\
  Department of Computer Science\\
  University of Southern California\\
  Los Angeles, CA \\
  \texttt{aferber@usc.edu} \\
  \And
  Bryan Wilder \\
  School of Engineering and Applied Sciences\\
  Harvard University\\
  Cambridge, MA \\
  \texttt{bwilder@g.harvard.edu} \\
  \AND
  Bistra Dilkina\thanks{Corresponding author} \\
  Department of Computer Science\\
  University of Southern California\\
  Los Angeles, CA \\
  \texttt{dilkina@usc.edu} \\
  \And
  Milind Tambe \\
  School of Engineering and Applied Sciences\\
  Harvard University\\
  Cambridge, MA \\
  \texttt{milind\_tambe@harvard.edu}
}

\begin{document}

\maketitle

\begin{abstract}
Machine learning components commonly appear in larger decision-making pipelines; however, the model training process typically focuses only on a loss that measures accuracy between predicted values and ground truth values. Decision-focused learning explicitly integrates the downstream decision problem when training the predictive model, in order to optimize the quality of decisions induced by the predictions. It has been successfully applied to several limited combinatorial problem classes, such as those that can be expressed as linear programs (LP), and submodular optimization. However, these previous applications have uniformly focused on problems from specific classes with simple constraints. Here, we enable decision-focused learning for the broad class of problems that can be encoded as a Mixed Integer Linear Program (MIP), hence supporting arbitrary linear constraints over discrete and continuous variables. We show how to differentiate through a MIP by employing a cutting planes solution approach, which is an exact algorithm that iteratively adds constraints to a continuous relaxation of the problem until an integral solution is found. We evaluate our new end-to-end approach on several real world domains and show that it outperforms the standard two phase approaches that treat prediction and prescription separately, as well as a baseline approach of simply applying decision-focused learning to the LP relaxation of the MIP.

\end{abstract}

\section{Introduction}

We propose a method of \textit{training predictive models to directly optimize the quality of decisions that are made based on the model's predictions}. We are particularly interested in decision-making problems that take the form of \textit{Mixed Integer Programs (MIPs)} because they arise in settings as diverse as electrical grid load control \cite{mohsenian2010electrical}, RNA string prediction \cite{sato2011rna}, and many other industrial applications \cite{nemhauser2013mipimpact}. MIPs naturally arise in so many settings largely due to their flexibility, computational complexity (ability to capture NP-hard problems), and interpretability. In many practical situations it is often necessary to predict some component (e.g., the objective) of the MIP based on historical data, such as estimated demand \cite{eoin2015citibike}, price forecasts \cite{delarue2010unitcommitment}, or patient readmission rate \cite{chan2012readmission}. However, these predictive models are often trained without regard for the downstream optimization problem. A standard loss function may place emphasis on correctly predicting outliers or small values because they greatly impact the loss function. However, in practice it may be the case that decisions will never be made on outliers, and so model capacity may be going towards adequately predicting values for a region of the feature space that is readily disregarded in the decision problem. Furthermore, practitioners may want to enforce that the outputs of the predictions meet semantically meaningful objectives such as ensuring predictions result in making   fair decisions downstream \cite{benabbou2018diversity, trilling2006nurse, warner1976fairnessnurses}. 

Machine learning components commonly appear in larger decision-making pipelines; however, the model training process typically focuses only on a loss that measures accuracy between predicted values and ground truth values. Decision-focused learning, introduced for a financial criterion in \cite{bengio1997financial} and extended to the more general quadratic programs in \cite{amos2017optnet} and linear programs in \cite{wilder2018melding},  explicitly integrates the downstream decision problem when training the predictive model, in order to optimize the quality of decisions induced by the predictions. 
In the commonly used gradient-based predictive models, the central challenge is in passing gradients back to give the predictive model an indication of how it should shift its weights in order to improve decision quality of the resulting optimal solution. The \textit{discrete} and \textit{discontinuous} solution space that makes the MIP so widely applicable also prevents us from easily differentiating through it, as has been done for embedding continuous optimization problems in neural networks \cite{amos2017optnet, wilder2018melding, de2018diffphys}. Our approach to computing gradients relies on the fact that we can \textit{algorithmically generate an exact continuous surrogate} for the original discrete optimization problem. We employ previous work on cutting plane approaches, which exactly solve a MIP by iteratively solving a continuous relaxation and cutting off the discovered integer infeasible solution until an integer feasible solution is found \cite{gomory1960cuts}. The final continuous, and convex, optimization problem can then be used for backpropagation by differentiating the KKT conditions \cite{karush1939minima} of the continuous surrogate, as has been suggested for convex problems \cite{amos2017optnet}. 
While pure cutting plane approaches are often slower than alternate branch-and-bound MIP solvers in practice \cite{Dash2014cuttingplanes}, we note that our approach only needs the cutting plane methodology for backpropagation during training. Indeed, at test time, we can make predictions and find optimal decisions based on those predictions using any state-of-the-art MIP solver, ensuring the running time is exactly the same were any other training method used.
Due to the computational complexity of computing the cutting planes for backpropagation, we also analyze a hybrid approach that stops cut generation after a fixed number of cuts have been generated, trading off the exactness of the differentiable solver with improved training runtime. An extreme version of this is when we disregard integrality requirements entirely and just use the LP relaxation of the problem for training.
Finally, we evaluate our decision-focused approaches against the baseline of a two-stage approach, that simply relies on training using a relevant classification or regression loss function. 

We demonstrate the effectiveness of our approach on two different real-world MIP problem domains concerning investment portfolio optimization and bipartite matching with diversity constraints, showing significant improvements in solution quality over the baseline.

\section{Problem description}

We consider problems that combine learning with an optimization problem that can be modeled as a mixed integer program (MIP). Specifically, each instance of the problem is a triple $(\phi, c, D)$, where $\phi$ is a feature vector, $c$ is the objective coefficient vector of a MIP, and $D$ represents additional known data that plays a role in the downstream optimization. In a MIP, $D$ will include the left hand and right hand constraint coefficients in each train instance $A,b$. If $c$ were known a priori, we could simply solve the MIP; however, we consider the setting where $c$ is unknown and must be estimated from $\phi$. We assume that we observe training instances $\{(\phi_1, c_1, D_1), \ldots, (\phi_m, c_m, D_m)\}$ drawn from some distribution. We will use this data to train a predictive model $f_\theta$ (where $\theta$ denotes the internal parameters of the model) which outputs an estimate $f_\theta(\phi) = \hat{c}$ on a test-time instance. The standard two stage approach in this setting is to train the machine learning model $f_\theta$ that minimizes a loss comparing predicted values $\hat{c}$ and ground truth values $c$. Our objective is to find model parameters $\theta$ which directly maximize the expected quality of MIP solution for $\hat{c}$, \emph{evaluated with respect to the (unknown) ground truth objective $c$}. We formalize this problem and our proposed approach below. 

\section{\MIPLayer: Encoding MIP in a Neural Network}

We formulate the MIP as a differentiable layer in a neural network which takes objective coefficients $\predc$ as input and outputs the optimal MIP solution. Formally, we consider the optimal solution $\optsol(\predc; A, b, \integral)$ of the MIP as a function of the input coefficients $\predc$ given linear constraints on the feasible region $Ax\leq b$ and the set of integral variables $\integral$. We write a functional form of the layer as:
\begin{equation}
\label{layer_formulation}
\optsol(\predc; A, b, \integral) =  
\begin{array}{ll@{}ll}
\text{argmin}_x  & \predc^Tx &\\
\text{subject to}& Ax & \leq b\\
& x_i \in \ZZ & \,\forall i \in \integral
\end{array}
\end{equation}

We can perform a forward pass given input objective coefficients, which are potentially outputs of a neural network, and feasibility parameters of the MIP using any solver. 

Standard practice in this setting is to first train a model $\predcmodel$ to predict the coefficients based on embeddings $\features_i$ of the different predicted components such that on average the model predictions $\predc=\predcmodel(\features)$ are not far away from the ground truth objective coefficients $\truec$. Then, decisions are made during deployment based on the predicted values by finding the optimal solution with respect to the predicted values $\optsol(\predc;A,b,\integral)$ based on \ref{layer_formulation}.

Forward propagation in this setting is straightforward; however, the highly nonconvex and discrete structure of the MIP, which enable it's flexibility, hinders straightforward gradient computation
for the backward pass. 

We propose an approach to compute the backward pass which relies on finding a continuous optimization problem which is equivalent to the original optimization problem at the optimal integral solution. In particular, we use a pure cutting plane approach which is guaranteed to give the optimal solution (at the expense of potentially generating exponentially many cutting planes along the way) \cite{bodur2017cutting, wolsey2014integer}.

The cutting plane approach iteratively solves the linear programming relaxation of the current problem. If the found solution is integral then the algorithm terminates since the found solution is both feasible to the original MIP and optimal for a relaxation of the original problem. Otherwise, a cut is generated which removes the discovered fractional solution and leaves all feasible integral solutions. Since the individual cuts do not remove any integral solutions, the  final LP retains all integral solutions. In the extreme case, assuming the feasible region is bounded, we could describe the convex hull of the integral solutions yielding a linear program equivalent to the original MIP, thus ensuring that all potentially optimal integral solutions lie on extreme points of the feasible region. The simplex algorithm, a practically efficient LP solver, finds the optimal extreme point
. In practice, finding the convex hull is intractable but we can obtain cutting planes that yield an exact integer solution via Gomory cuts or other globally valid cuts \cite{gomory1960cuts,balas1996gomory}.

We can then consider that we have generated cuts $S x\leq t$ and write out the following equivalent linear program:
\begin{equation}
\label{continuous_opt}
\begin{array}{ll@{}ll}
\text{minimize}_x  & \predc^Tx &\\
\text{subject to}& Ax & \leq b\\
& Sx & \leq t
\end{array}
\end{equation}
Given this continuous optimization problem, we can now find the gradient of the optimal solution with respect to the input parameters by differentiating through the KKT conditions which are necessary and sufficient for the optimal solution of \ref{continuous_opt} (and hence also of the original MIP). This is done via the quadratic smoothing approach proposed in \cite{wilder2018melding} for linear programs based on differentiating the KKT conditions of a quadratic program as shown in \cite{amos2017optnet}.

\section{Decision-Focused Learning with \MIPLayer{}}
Training to minimize the error between predicted and ground truth values $\predc$ and $\truec$ can yield decisions that have poor performance in practice in that the standard error metrics like mean squared error or crossentropy may not be directly aligned with the decision quality of the solutions obtained.
To remedy this, we can train the predictive component to perform well by defining the loss to be the solution quality of the solution given the predicted coefficients. In other words, we can use the \MIPLayer{} formulation to train the model to directly minimize the deployment objective. Using a neural network parametrized by $\theta$, $f_\theta$ to predict objective coefficients based on embeddings $\features$ of the decision variables, we compute one forward pass as shown in \ref{decision-focused-forward} utilizing the cutting plane solver to generate the LP corresponding to the original MIP.
\begin{subequations}
\label{decision-focused-forward}
\begin{align}
\predc &:= f_\theta(\features) \label{eqn:nnpred}\\
\hat{x} &:= \optsol(\predc; A,b, \integral) & \text{via \ref{layer_formulation}} \label{eqn:optpred}\\
loss(\predc, \truec) &:= \truec^T \hat{x} \label{eqn:dotprod}
\end{align}
\end{subequations}
Since the dot product in \ref{eqn:dotprod} and prediction of the neural network in \ref{eqn:nnpred} are differentiable functions of their inputs, the parameters of the neural network predictor $\theta$ can be trained via backpropagation using KKT conditions of the computed surrogate LP found in \ref{continuous_opt}, relying on a small quadratic regularization term for the LP proposed in \cite{wilder2018melding} to enforce strong convexity and perform backpropagation through \ref{eqn:optpred}.

\section{Empirical Evaluation}
We instantiate \MIPLayer{} for a range of tasks which require predicting and optimization with the overall goal of improving the objectives upon deployment of the recommendations. Specifically, we run experiments on \textit{combinatorial portfolio optimization} and \textit{diverse bipartite matching}. The portfolio optimization setting accounts for various combinatorial constraints enforcing small cardinality of the assets in the portfolio and limiting rebalancing transactions to maximize the overall predicted return of the portfolio. Diverse bipartite matching enforces diversity in the types of pairs that are recommended to maximize an overall predicted utility of the suggested pairing. In each setting, the predictive problem is nontrivial since the features do not contain much signal for the predictive task. However, we demonstrate that in these settings, a predictive model can perform well when given the specific task of ensuring that the predictions yield high-quality decisions.
\subsection{Decision-Making Settings}
\textbf{Combinatorial Portfolio Optimization} arises in several financial applications, with convex variants motivating early work in developing loss functions directly tied to decision quality \cite{bengio1997financial}. However, the previously considered Markowitz portfolio optimization problem \cite{markowitz1952portfolio} is unable to capture discrete decisions which are desired or necessary to meet combinatorial deployment requirements as done in \cite{bertsimas1999portfolio}, where the authors use MIP to maximize percentage increase while limiting fixed costs incurred on trades, and limiting position changes for individual industries. We use a slightly modified version of the MIP from \cite{bertsimas1999portfolio} which we specify fully in the appendix. The resulting formulation leverages the real-world modeling flexibility that MIPs afford: including big-M constraints, piecewise linear function modeling, and logical implication.

In the combined prediction and optimization problem, the next period's percent increase in different assets (i.e., the objective) are unknown and must be learned from historical data. We gather features from the Quandl WIKI and FSE datasets \cite{quandl} (which aggregate indicators for each company). We evaluate on the SP500, a collection of the 505 largest companies representing the American market, and the DAX, a set of 30 large German companies. We split the data temporally into train, test, and validation sets. More details about the data collection may be found in the appendix.

\paragraph{Diverse Bipartite Matching} 
Bipartite matching is used in many applications, where the benefit of a particular matching is often estimated from data. Without additional constraints, bipartite matching can be formulated as an LP; this formulation has previously been used for decision-focused learning \cite{wilder2018melding} since the LP relaxation is exact. In practice though, matching problems are often subject to additional constraints. For instance, finding fair and high-quality housing allocations or kidney matching with additional contingency plans \cite{benabbou2018diversity, dickerson2016kidney} require additional decision variables and constraints on the solutions which make the integer problem NP-Hard and thus not solvable using a polynomially-sized LP, unless P=NP. 


We use the problem of bipartite matching with added diversity constraints, which enforce a maximum and minimum bound on the percent of edges selected with a specified property. We use the experimental setup of \cite{wilder2018melding}, who did not include diversity constraints. Specifically, the matching instances are constructed from the CORA citation network \cite{sen2008cora} by partitioning the graph into 27 instances on disjoint sets of nodes (split into train, test and validation). Diversity constraints are added to enforce that at least some percent of edges selected are between papers in the same field and some percent are between papers in different fields.

\subsection{Baselines}

\textbf{Two-Stage: } We compare against the standard predict-then-optimize approach which treats prediction and optimization components separately. The predictive component is trained to minimize a standard loss between predicted objective coefficients and the ground truth (e.g., mean squared error or cross-entropy). Afterwards, we solve the MIP to optimality using the predicted coefficients. 

\textbf{RootLP: } Next, we compare against an alternate decision-focused learning method from \cite{wilder2018melding}, which uses only the naive LP relaxation of the MIP (disregarding integrality constraints). The predictive model is trained using the LP relaxation, and at test time we solve the true MIP using the predicted objective coefficients to obtain an integral decision. This tests the impact of our cutting plane method, which allows us to fully account for combinatorial constraints in \MIPLayer{}'s gradients.

\textbf{\MIPLayer{} - k cuts: } Given that the cut generation process is time consuming and must be done for each forward pass, we examine the tradeoff of decision quality when limiting the generated cuts to a fixed number $k$, and stopping cut generation after the first $k$ cuts have been generated. This essentially attempts to solve the original generally NP-Hard problem in polynomial time since each LP is solvable in polynomial time and we generate a constant number of cuts. We experiment with two settings ($k=100$ and $k=1000$) to determine how the exactness of the cut generation process impacts the decision quality at test-time. Note that in the case where $k=0$ cuts are generated, this method is equivalent to the RootLP approach where no cuts are generated.

\subsection{Metrics}
We evaluate the deployment quality of \MIPLayer{}'s outputs as well as the similarity of the predicted MIP inputs to the ground truth objective coefficients. Where applicable, we provide $95\%$ confidence half-widths to indicate our confidence in the mean evaluation. For comparative metrics, we use a one-sided paired t-test with significance level of 0.05 to indicate whether we can reject the hypothesis that observed improvement is due to random chance.

\textbf{Decision quality:}
The decision quality of a given model's outputs is determined by the objective value after deployment of the model. A trained ML model predicts the objective coefficients of the particular decision problem, and CPLEX, a fast commercial optimization solver, gives an optimal solution based on those predictions. We evaluate the proposed solution based on the objective value with respect to realized ground truth objective coefficients. For portfolio optimization, the decision quality corresponds to the percentage change in the portfolio value from one time period to the next so the $2.79\%$ that \MIPLayer{}-Exact achieves in \ref{tab:decision-quality} indicates that with this method we increase the portfolio value by $2.79\%$ every month on average. In bipartite matching, our goal is to maximize the number of successful matches and so the solution quality, as in previous work \cite{wilder2018melding}, corresponds to the average number of successful matches in the given instance. Top entries are bolded along with any method whose confidence interval overlaps with that of the best entry.

\textbf{ML performance:}
We show the predictive performance for the given models in \ref{tab:ml-performance}. For portfolio optimization (which is a regression problem), we report MSE, while for bipartite matching (a classification problem), we report Cross-Entropy (CE) and AUC. We also visualize each model's predictions in \ref{fig:scatter} to better understand performance improvement. 



\subsection{Experimental Results}\textbf{Setup: }Experiments are run on a cluster of five identical 32-core machines with Intel 2.1 GHz processors and 264 GB of memory. We use the C API of IBM ILOG CPLEX 12.6.1 to generate and record cutting planes during training. Neural networks are trained and tested with the PyTorch Python API \cite{paszke2017pytorch} and evaluated with the CPLEX's Python API. We average over 5 training and testing iterations per problem setting with different seeds to evaluate the given approaches. This results in 180 portfolio optimization instances, and 135 diverse bipartite matching instances.

\textbf{Predictive architectures: }The neural networks for the portfolio optimization problems consist of two fully-connected layers with 100 nodes each, represented as a fully-connected layer, followed by batch normalization \cite{ioffe2015batchnorm}, LeakyRelu as a nonlinearity \cite{maas2013leakyrelu}, and dropout with probability 0.5 as proposed and suggested in \cite{srivastava2014dropout}. As portfolio optimization is built on a regression problem, we add a linear layer on the output of the predictive component. The architecture for matching is similar although it has only one layer and uses a sigmoid activation function on the output of the predictive component since the task is edge classification. These architectures were selected from grid search based on the task-based validation loss of the standard two stage model. The grid search was done considering models with between one and three layers, ReLU, LeakyReLU, and Sigmoid activation functions, and either 100 or 200 nodes per layer. We additionally, ran experiments disregarding dropout and batch normalization and found that without either the networks were unable to generalize. The models are trained with the Adam optimizer \cite{kingma2014adam} with learning rate 0.01 and l2 normalization coefficient of 0.01 which were selected based on TwoStage decision quality.

\begin{table}
\centering
\caption{Decision quality comparison of portfolio optimization (monthly percentage increase), and bipartite matching (number of pairs successfully matched). \MIPLayer{} gives 2x monthly returns on SP500 and 8x on DAX compared to TwoStage, and successfully matches 40.3\% more edges.}
\label{tab:decision-quality} 
\begin{tabular}{rccc}
    \toprule
    &SP500 & DAX & Matching\\
    \midrule
    MIPaaL-Exact          & \textbf{2.79 $\pm$ 0.17}   & \textbf{5.70 $\pm$ 0.68} & \textbf{4.80 $\pm$ 0.71}              \\
MIPaaL-1000           & \textbf{2.60 $\pm$ 0.16}   & \textbf{4.39 $\pm$ 0.66} & \textbf{3.45 $\pm$ 0.71}              \\
MIPaaL-100            & 1.25 $\pm$ 0.14            & 0.35 $\pm$ 0.63          & 2.57 $\pm$ 0.54                       \\
RootLP                & 1.97 $\pm$ 0.17            & -1.97 $\pm$ 0.69         & 3.17 $\pm$ 0.60                       \\
TwoStage              & 1.19 $\pm$ 0.15            & 0.70 $\pm$ 1.46          & 3.42 $\pm$ 0.78                     \\
    \bottomrule
  \end{tabular}
\end{table}

\textbf{Results}: As shown in \ref{tab:decision-quality}, the exact \MIPLayer{}  gives uniformly higher decision quality than baseline methods, with \textit{more than 2x the average return of the TwoStage or RootLP methods} for portfolio optimization and \textit{40.3\% more successful matches} for biparite matching. These results drive home the importance of integrating the full combinatorial problem into training, as enabled by \MIPLayer{}. Comparing between versions of \MIPLayer{}, we find that stopping cut generation at $k = 1000$ results in competitive performance in an aggregate sense with the exact \MIPLayer{} layer. Interestingly, stopping cut generation too early at $k = 100$ underperforms RootLP on SP500 and Matching.

\begin{table}
\centering
\caption{Decision quality per instance win / loss percentages, with * indicating decision performance of winner is statistically significant. \MIPLayer{} reliably improves or ties in these settings.} \label{tab:win-tie-loss} 
\begin{tabular}{crcccc}
\toprule
                                                                            &              & MIPaaL-1000                     & MIPaaL-100                      & RootLP                          & TwoStage   \\ \midrule
                                                                            & MIPaaL-Exact & 57.2 / 42.3 * & 87.2 / 12.8 * & 82.8 / 17.2 * & 90.6 / 9.4 * \\ 
\multirow{2}{*}{SP500}                                                      & MIPaaL-1000  &            & 84.4 / 15.6 * & 80.0 / 20.0 * & 85.0  /  15.0 * \\ 
                                                                            & MIPaaL-100   &            &            & 27.8 / 72.2 * & 52.8 / 47.2    \\ 
                                                                            & RootLP       &            &            &            & 69.4 / 30.6 * \\ 
                                                                            \midrule
                                                                            & MIPaaL-Exact & 66.7 / 33.3 * & 86.6 / 13.4 * & 88.9 / 11.1 * & 84.4 / 15.6 * \\
\multirow{2}{*}{DAX}                                                        & MIPaaL-1000  &            & 81.1 / 18.9 * & 85.6 / 14.4 * & 73.9 / 26.1 * \\ 
                                                                            & MIPaaL-100   &            &            & 69.4 / 30.6 * & 43.3 / 56.7   \\ 
                                                                            & RootLP       &            &            &            & 28.3 / 71.7 * \\ \midrule
                                                                            & MIPaaL-Exact & 11.7 / 3.9 *   & 15.6 / 2.8 *   & 15.0 / 3.9 *   & 12.2 / 5.6 *  \\ 
\multirow{2}{*}{\begin{tabular}[c]{@{}c@{}}Diverse\\ Matching\end{tabular}} & MIPaaL-1000  &            & 11.7 / 3.9      & 9.4 / 8.3     & 10.0 / 7.2    \\ 
                                                                            & MIPaaL-100   &            &            & 3.9 / 11.1      & 5.0 / 8.9    \\ 
                                                                            & RootLP       &            &            &            & 7.2 / 8.3    \\ \midrule
\end{tabular}
\end{table}

To further tease out the benefit of using variants of \MIPLayer{} on a case-by-case basis, we compare decision quality in terms of win/loss rates in \ref{tab:win-tie-loss}. This table shows the percent of instances the given approach on the left won/lost to the approach on the top, matching instances based on optimization problem and seed. Additionally, we compare the means of these values using a one-sided t-test with significance level 0.05. Asterisks indicate statistically significant results.
Note that in portfolio optimization the allocated weights are somewhat continuous and so getting a tie is very rare. However, in the matching problem, we assign only integral pairs and get integral rewards if the match is successful or not so the objectives here are discrete and thus ties are more common. 

As is apparent in \ref{tab:win-tie-loss}, the \MIPLayer{} method tends to outperform the other methods on all problem settings on an individual instances basis. Also useful to note is that the RootLP method beats the \MIPLayer{}-100 approach in both the portfolio optimization settings, potentially indicating that limiting cut generation too much can be detrimental to the performance of the overall system. Additionally, in the diverse matching setting, the standard two stage approach does comparably to the $k$-cut decision-focused methods which stop cut generation early. This suggests that while the diverse matching problem is similar to the original bipartite matching LP, adding a few diversity constraints breaking the integrality property of the LP relaxation can remove gains made from decision-focused learning unless the exact \MIPLayer{} layer is used.

\begin{table}
\centering
\caption{ML performance. TwoStage generalizes on metrics used for training (MSE, CE), whereas \MIPLayer{} improves deployment loss and AUC. Edge prediction is hard as evidenced by AUCs near 0.5.} \label{tab:ml-performance} 
\begin{tabular}{rcccc}
\toprule
& SP500 & DAX & \multicolumn{2}{c}{Matching}\\
\cmidrule(r){2-2} \cmidrule(r){3-3} \cmidrule(r){4-5}
 & MSE & MSE & AUC & CE \\
\midrule
MIPaaL-Exact          & 0.215 $\pm$ 0.043          & 0.126 $\pm$ 0.017          & \textbf{0.535 $\pm$ 0.004} & 0.658 $\pm$ 0.009          \\
MIPaaL-1000           & 0.117 $\pm$ 0.020          & 0.349 $\pm$ 0.010          & 0.506 $\pm$ 0.007          & 0.614 $\pm$ 0.010          \\
MIPaaL-100            & 0.983 $\pm$ 0.089          & 0.989 $\pm$ 0.060          & 0.503 $\pm$ 0.004          & 0.543 $\pm$ 0.013          \\
RootLP                & 0.705 $\pm$ 0.178          & 1.055 $\pm$ 0.137          & 0.513 $\pm$ 0.001          & 0.493 $\pm$ 0.007          \\
TwoStage              & \textbf{0.086 $\pm$ 0.017} & \textbf{0.022 $\pm$ 0.066} & 0.514 $\pm$ 0.005          & \textbf{0.392 $\pm$ 0.004} \\ \hline
\end{tabular}
\end{table}

Looking at the \textit{machine learning metrics} in \ref{tab:ml-performance}, we notice that the final machine learning performance of the decision-focused methods vary widely. In particular, the testing mean squared error for the portfolio optimization problems is quite high compared to the two stage approach. This mismatch between the MSE and decision quality exemplifies the need for training with the task in mind in that even though the \MIPLayer{} model has worse MSE than TwoStage in both portfolio optimization settings, it results in much higher-return decisions. 


\begin{figure}
\centering

\begin{minipage}{0.49\linewidth}
\includegraphics[width=\linewidth]{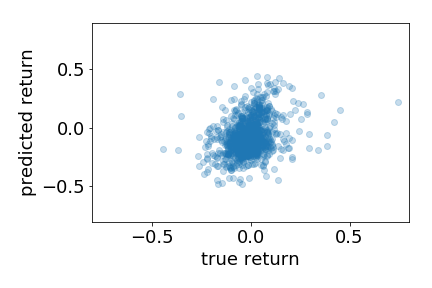}
\subcaption{\MIPLayer{} predictions on DAX}\label{fig:mipdax}
\end{minipage}
\begin{minipage}{0.49\linewidth}
\includegraphics[width=\linewidth]{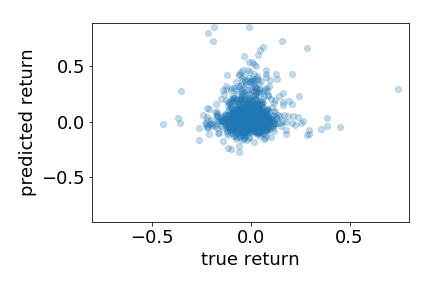}
\subcaption{TwoStage predictions on DAX}\label{fig:2stagedax}
\end{minipage}

\begin{minipage}{0.49\linewidth}
\includegraphics[width=\linewidth]{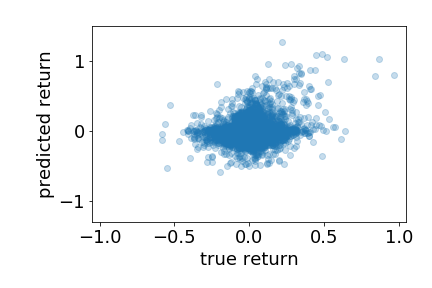}
\subcaption{\MIPLayer{} predictions on SP500}\label{fig:mipsp500}
\end{minipage}
\begin{minipage}{0.49\linewidth}
\includegraphics[width=\linewidth]{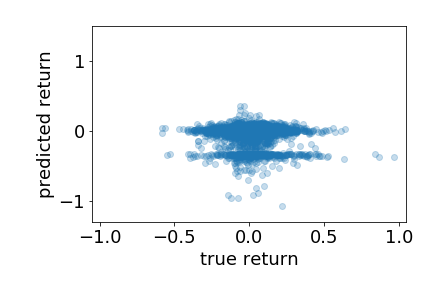}
\subcaption{TwoStage predictions on SP500}\label{fig:2stagesp500}
\end{minipage}
\caption{Scatter plots of predicted coefficients vs ground truth returns.}\label{fig:scatter}

\end{figure}

\ref{fig:scatter} gives a better understanding of how this occurs. Examining \ref{fig:mipdax} and \ref{fig:mipsp500}, we notice that even though the model learned with \MIPLayer{} does poorly in general, it focuses in on identifying points that yield very high return while disregarding elements of mediocre return, leaving them in the center. The two-stage approach seen in \ref{fig:2stagedax} and \ref{fig:2stagesp500} is focused more heavily on predicting many of the values somewhat well, disregarding that incorrectly predicting an asset to have high return may mean that it incorrectly ends up in the portfolio. Looking comparatively at the predictions of \MIPLayer{} in \ref{fig:mipdax} and TwoStage in \ref{fig:2stagedax}, we can see that TwoStage predicts a large column of assets to have high returns when in practice they have relatively low or even negative returns. On the other hand, \MIPLayer{} predicts relatively high returns correctly while mostly disregarding assets with lower returns. This is even more apparent looking at \ref{fig:mipsp500} and \ref{fig:2stagesp500} as \MIPLayer{} correctly predicts values for very profitable assets while TwoStage predicts relatively low or negative returns for most assets.

In terms of the matching problem, we see that even though TwoStage has better Cross-Entropy loss at test time, as it was trained with that specific classification loss in mind, it lacks in AUC which both corresponds to the findings in previous work \cite{wilder2018melding}, and indicates that the predictions learned by \MIPLayer{} may sometimes also be accurate in a traditional sense.

\begin{table}
\centering
\caption{Transfer learning across data distributions: trained on SP-$\text{30}^a$} \label{tab:transfer-learning} 
\begin{tabular}{rccccc}
\toprule
&\multicolumn{2}{c}{Decision quality} &  \multicolumn{2}{c}{MSE}\\
& SP-$\text{30}^b$ & DAX & SP-$\text{30}^b$ & DAX\\
\cmidrule(r){2-3} \cmidrule(r){4-5}
\MIPLayer{} & \textbf{2.02  $\pm$ 0.48}  & \textbf{2.77 $\pm$ 0.40} & 4.81 $\pm$ 8.59 & 4.59 $\pm$ 8.80\\
RootLP & \textbf{1.81  $\pm$ 0.44}  & 1.74 $\pm$ 0.43 & 5.14 $\pm$ 1.02 & 5.39 $\pm$ 1.04\\
TwoStage & 0.71  $\pm$ 0.04  & 0.82 $\pm$ 0.54 & \textbf{0.079 $\pm$ 0.052} & \textbf{0.065 $\pm$ 0.032} \\
\bottomrule
\end{tabular}
\end{table}

\begin{table}
\centering
\caption{Transfer learning across problem sizes: trained on SP-$\text{30}^a$. Decision quality, monthly rate of return, on different problem sizes. \MIPLayer{} ensures that the predictions perform well in multiple deployment settings whereas the TwoStage approach is unable to extract relevant information for larger scale problems.} \label{tab:transfer-learning-sp500} 
\begin{tabular}{rcccc}
\toprule
& SP-50 & SP-100 & SP-200 & SP500\\
\midrule
\MIPLayer{} & \textbf{1.93 $\pm$ 0.13} & \textbf{2.27 $\pm$ 0.11} & \textbf{2.17 $\pm$ 0.48} & \textbf{2.26 $\pm$ 0.37} \\
RootLP      & 1.50 $\pm$ 0.09 & 1.58 $\pm$ 0.08 & \textbf{1.82 $\pm$ 0.41} & \textbf{1.90 $\pm$ 0.29} \\
TwoStage    & 1.58 $\pm$ 0.13 & 1.22 $\pm$ 0.09 & 1.50 $\pm$ 0.58 & 1.11 $\pm$ 0.35 \\
\bottomrule
\end{tabular}
\end{table}

\begin{table}
\centering
\caption{Transfer learning MSE results on different problem sizes: trained on SP-$\text{30}^a$. The TwoStage approach clearly does better in approximating the distribution of objective coefficients even when the data distribution changes. However, the improvement in MSE does not ensure good deployment quality as show in \ref{tab:transfer-learning-sp500}} \label{tab:transfer-learning-ml} 
\begin{tabular}{rcccc}
\toprule
& SP-50 & SP-100 & SP-200 & SP500\\
\midrule
\MIPLayer{} & 5.421 $\pm$ 3.163  & 5.422 $\pm$ 2.368    & 5.254 $\pm$ 1.834    & 5.431 $\pm$ 1.670    \\
RootLP      & 4.725 $\pm$ 3.169  & 4.876 $\pm$ 2.579    & 4.807 $\pm$ 1.906    & 4.834 $\pm$ 1.556    \\
TwoStage    & \textbf{0.078 $\pm$ 0.019}  & \textbf{0.074 $\pm$ 0.012}    & \textbf{0.078 $\pm$ 0.011}    & \textbf{0.076 $\pm$ 0.011}    \\

\bottomrule
\end{tabular}
\end{table}

\textbf{Transfer learning:}
We evaluate \MIPLayer{}, RootLP, and TwoStage on two transfer learning tasks for portfolio optimization. Here, models are trained on 30 assets randomly drawn from SP500 (SP-$\text{30}^a$), with data from January 2005 - December 2010. We then evaluate each model on data from December
2013 to November 2016 based on 1) SP-$\text{30}^b$, a set of 30 randomly drawn assets from the SP500, disjoint from SP-$\text{30}^a$, and 2) the DAX, a separate index comprising 30 companies from a different country.
The transfer learning results in \ref{tab:transfer-learning} demonstrate that \MIPLayer{} not only performs well across time periods, but generalizes to unseen assets as well as unseen countries. On SP-$\text{30}^b$ \MIPLayer{} gives more than double the improvement in the average rate of return over the standard TwoStage approach, and a 59\% improvement over RootLP, indicating  \MIPLayer{}'s good generalization performance. Furthermore, while the transferred \MIPLayer{} applied to DAX doesn't beat \MIPLayer{} trained on the same task, shown in \ref{tab:decision-quality}, it improves performance over the RootLP and TwoStage in both settings, showing that \MIPLayer{} learns a useful model for portfolio optimization as a whole rather than approximating the training data distribution. Lastly, \MIPLayer{}'s performance improvement over TwoStage occurs despite a higher MSE, highlighting the importance of the decision-focused loss function.

\textbf{Transfer Learning: Problem Size Generalization}
We evaluate the abilit of \MIPLayer{} to perform well when predictions are required in differently sized settings. In particular, we evaluate when a model is trained to perform well on the portfolio optimization with 30 assets, but used in deployment on a larger number of assets. SP-50, SP-100, and SP-200 are instances drawn on 3 sets of assets which are all disjoint from both SP-$\text{30}^a$ and SP-$\text{30}^b$, with 50, 100, and 200 assets each. Again, we only evaluate on testing time periods to prevent data leakage and present results on the monthly rates of return in \ref{tab:transfer-learning-sp500} and MSE values in \ref{tab:transfer-learning-ml}.

As shown in \ref{tab:transfer-learning-sp500}, decision-focused approaches perform well when the predictions are used to find the optimal MIP solution. In this case the performance improvement is not as drastic as when the MIP is trained directly for the task in question. However, these results demonstrate that \MIPLayer{} is able to efficiently extract information relevant to the general task of portfolio optimization, regardless of the set or number of assets in question. 

\section{Related Work}

The interaction of machine learning and optimization has recently provided a new set of tools for efficiently solving a wide variety of problems. Recent work has framed traditionally heuristic components of optimization algorithms as machine learning tasks such as learning in branch and bound for MIPs \cite{alvarez2014learning, khalil2016branching, khalil2017heuristics, lodi2017learning, he2014nodeselection, horst2012nodeselection, balcan2018l2b, bengio2018ml4opttour}. These approaches work inside an exact solver when the objective is known and not predicted from data. Deep reinforcement learning methods have also been used to generate high-quality data-driven solutions to hard problems such as graph optimization \cite{graphcombopt}, vehicle routing \cite{vinyals2015pointer, kool2018attention, nazari2018vrp}, and realtime patrol planning in security games \cite{yu2018rlgsg}. These problems also take as input different known features such as the true objective coefficients and are mainly used to help quickly generate high-quality solutions with a known input, which would be computationally intractable or take an unknown amount of time for an exact solver. 

Several approaches have been proposed which embed optimization components in neural networks. This includes specific problems such as Markowitz portfolio optimization \cite{bengio1997financial} or physically feasible state transitions \cite{de2018diffphys}, as well as larger classes which exhibit properties like convexity or submodularity. Problem classes used in end-to-end training include quadratic programs \cite{amos2017optnet}, linear programs \cite{wilder2018melding}, and zero-sum games \cite{ling2018game, perrault2019decision}. In addition, a solution is proposed for solving submodular optimization problems in \cite{wilder2018melding}. In \cite{elmachtoub2017spo}, the authors create a surrogate loss function and connect it to a decision-focused regret bound. Additionally, in \cite{NIPS2009_3686} the authors instantiate end-to-end learning with linera models predicting components of a quadratic optimization function. Lastly, \cite{wang2019satnet} incorporates a differentiable semidefinite programming problem as a relaxation for MAXSAT instances. To our knowledge, \MIPLayer{} is the first approach for imbuing neural networks with the highly flexible Mixed-Integer Program, a widely-used class of potentially inapproximable NP-Hard optimization problems, while providing exact feedback on the decision quality.

\section{Conclusion}

We propose \MIPLayer{}, a principled method for incorporating Mixed Integer Programs as a differentiable layer in neural networks. We approach the task of differentiating with respect to this flexible, discrete, and potentially inapproximable problem by algorithmically generating an equivalent continuous optimization problem via cutting planes. We instantiate our proposed approach for decision-focused learning wherein a predictive model is trained with a loss function that directly corresponds to the quality of the decisions made based on the predictions. \MIPLayer{} is evaluated on two settings of portfolio optimization and one setting of bipartite matching with additional diversity constraints, which contain many modeling techniques widely used in combinatorial optimization that make the problem more complex but also more realistic. We demonstrate empirically that \MIPLayer{} is able to outperform the standard approach of decoupling the prediction and decision components, as well as an approach of just using a continuous relaxation of the original combinatorial optimization problem. To better understand the impact of the cutting plane technique, we explore hybrid strategies that stop the cutting plane generation early. Ultimately, we find empirically that our approach can give high-quality solutions in the investigated settings.

\newpage
\bibliographystyle{plain}

\appendix
\section{Portfolio Optimization Specifics}

\subsection{Data specification}
We simulate our approach on this problem setting we use historical price and volume data downloaded from the Quandl WIKI dataset \cite{quandl}. Our goal is to generate monthly portfolios of stocks in a given market index which targets a portfolio of these stocks weighted based on market capitalization, assuming that we start out with a portfolio that is weighted by market capitalization from the previous month. We train our model on data collected from January 2005 to December 2010, validate on data collected from January 2011 to November 2013 and test our model based on data from December 2013 to November 2016 resulting in 72 time periods used for training, 35 for validation and 36 for testing. We split the data temporally to ensure that data from the future isn't used to inform predictions in the current time period. The 11 features used are made up of historical price averages and rates of return over different time horizons, which are commonly used indicators for simple trading strategies. Furthermore, to evaluate the generality of the approach we evaluate on two market indices: the SP500 and DAX. The SP500 is an index of 505 large companies in the united states which are widely representative of the US trading market. The DAX is composed of 30 large German companies and  is much smaller than the SP500, being somewhat less representative of its respective market as a whole, but still representative of the central publicly traded companies in terms of trading volume and amount invested in those companies.\\

We use the following 11 indicators commonly used in simple momentum-based and mean-reversion trading strategies \cite{conrad1998trading}:\\
percentage increase of price from previous 5 time periods\\
percentage increase of price from previous year\\
percentage increase of price from previous quarter\\
mean of price percentage increase over previous year\\
variance of price percentage increase over previous year\\
mean of price percentage increase over previous quarter\\
variance of price percentage increase over previous quarter\\

We randomly sample 60 indices from the SP500 for SP-$\text{30}^a$ and SP-$\text{30}^b$, these indices are:\\
\textbf{SP-$\text{30}^a$} = LRCX, PHM, WY, SPG, EMN, CME, AEP, F, CAG, FISV, WBA, XOM
, NVDA, ETN, MDT, FL, HBAN, FFIV, BLK, IPG, EXPD, IRM, PH
, DLTR, COST, NBL, INCY, CSX]\\
 \textbf{SP-$\text{30}^b$} = [DVA, DE, BAC, KLAC, ADBE, FIS, IT, KR, FMC, HOG, SHW, RE
, ETR, BK, ACN, NWL, ESS, VMC, C, EW, IR, SWKS, SNPS, ARE
, SCHW, WEC, IVZ, SLB\\

Indices for the SP-50, SP-100, and SP-200 are:\\
\textbf{SP-50} = VLO, EIX, RSG, PKG, CMI, AMZN, CHRW, ATVI, ADS, RHI, DVN
, WMB, ILMN, LOW, SIVB, T, NTRS, HCP, ALXN, CPB, MRO, MU, CB
, HES, APA, VFC, CHD, CMCSA, ALK, PBCT, BBY, MNST, UDR, CTXS
, AZO, XRX, SBAC, M, MTD, COP, UTX, KO, MCO, TGT, CELG, HSIC
, WYNN, YUM, ECL, ABT\\

\textbf{SP-100} = RTN, K, ROST, WAB, MMM, WMT, JPM, MMC, AMD, BA, NEM, JBHT
, STI, ITW, NKTR, ETFC, PNW, HSY, EL, MLM, UPS, FRT, ES, ROL
, CMS, MAC, PVH, TSN, ANSS, SJM, DISH, LUV, MOS, TTWO, SEE
, PKI, BAX, AMG, EOG, AMGN, CCI, EA, AGN, HP, HAL, AXP, SRE
, INTU, AOS, APH, ROK, ABMD, CINF, UNM, ZION, HRL, ADSK, MSFT
, JNJ, EBAY, ADI, EXC, CVS, PEG, CVX, PNR, PAYX, CPRT, DHR
, JCI, SYK, HST, MO, INTC, GD, MCHP, PFG, PG, QCOM, STX, HAS
, WM, CL, CDNS, BF\_B, HIG, REGN, COG, RL, GPS, APD, APC, GPC
, TSCO, JKHY, FAST, NOV, TMK, AVY, PGR\\

\textbf{SP-200} = SYMC, WDC, CSCO, STZ, NUE, MAA, ARNC, DGX, EQIX, PWR, ED
, LLY, AIV, ISRG, FITB, EQR, BXP, ORLY, MHK, JEC, UNP, TXT
, IDXX, WHR, OMC, KEY, NI, FLIR, SNA, MCK, PXD, MCD, XEC, LEG
, KSU, TSS, IP, AVB, RF, GPN, AFL, UNH, DTE, NEE, ZBH, NDAQ
, DRE, MSI, VRSN, A, XEL, MAS, LKQ, FDX, LMT, WFC, NRG, RHT
, COF, RMD, SLG, TROW, ANTM, BRK\_B, PSA, RJF, BSX, AJG, FLR
, BLL, AES, ORCL, MAR, HD, GILD, RCL, TIF, PRU, SWK, TMO, NOC
, NKE, IFF, DHI, STT, AMT, TJX, PRGO, GS, NFLX, GLW, PFE, DOV
, OKE, PLD, VZ, LNC, PNC, CTSH, JNPR, LEN, ABC, CI, CCL, LH
, FCX, FLS, PCAR, HRS, EMR, GRMN, AKAM, BBT, PEP, OXY, KSS, MS
, URI, AEE, NSC, CAH, VTR, O, USB, SYY, FE, TXN, BDX, BWA
, CMA, COO, NTAP, SBUX, GWW, CNP, HOLX, CNC, D, TAP, MET, AAP
, TFX, IBM, XLNX, LLL, AMAT, HUM, AON, DIS, KMB, BEN, GIS, ADM
, MXIM, PPL, HPQ, ALL, VRTX, L, KIM, DOW, HON, DRI, AAPL, XRAY
, MKC, MRK, HFC, FOX, CTAS, REG, HRB, KMX, SO, MYL, CLX, CERN
, ALB, BIIB, LNT, BMY, MAT, DUK, MTB, AME, LB, GE, VNO, ROP
, EFX, AIG, JWN, CAT, UHS, WAT, ADP, ATO, VAR, FOXA, MDLZ

\subsection{Optimization model} 
We specify the optimization model used for the portfolio optimization task.\\

For our applications, we set the ticket budget $\ticketmax$, and $\namemax$ to be half of the number securities considered, 200 for SP500, and 15 for DAX and SP-$30$. We set the sector budget $\sectormax$ to be 0.1, in that for any given sector we can change at most 10\% of the portfolio weight into or out of that sector. Overall we found that this gave us a balance of reasonable problems that weren't trivial to solve but still had a non-empty feasible region.\\

\textbf{Given:} Set of $n$ assets $i=1,\dots,n$, set of sectors $S$ represented as a partitioning of the $n$ assets\\
Ticket (trading) limit $\ticketmax$\\
Name (unique item changes) limit $\namemax$\\
Sector deviation limit $\sectormax$\\
For each asset $i=1,\dots,n$:\\
Expected returns $\alpha_i$\\
Trading volume $\volume$\\
Indicator whether asset $i$ belongs to sector $s$\\ $M_s(i)\equiv i\in s$\\
Initial portfolio asset weight $w_0(i)$\\
Target portfolio asset weight $w_t(i)$\\

\textbf{Goal:} find final portfolio weights $w_f(i)$ which is close to the target portfolio $w_t$(i) in terms of weight, doesn't incur too much cost from $w_0(i)$ to execute, respects budget constraints in terms of sector exposure, cardinality, and trading limits, and ensuring all weights add up to 1\\

\textbf{Decision variables:} 
final weight in asset $i$ $w_f(i) \in [0,1]  \forall i = 1,\dots, n$ \\
auxiliary ticket-counting variables $z_1(i), z_2(i), z_3(i)$\\
auxiliary indicator of whether asset $i$ is used $y_\text{names}(i)$\\
auxiliary indicator of whether asset $i$ changes weight in the portfolio $y_\text{tickets}$\\
auxiliary variable $f(i)$ representing absolute value of weight change from original, keeping track of tickets bought\\
auxiliary variable $y(i)$ representing absolute value of deviation from target\\
auxiliary variable $x(s)$ representing absolute value of sector weight change\\
auxiliary variables $z_1(i), z_2(i), z_3(i)$ to put weights in three different compartments of a piecewise linear function related to volume.

\textbf{Formulation:}
\begin{equation}
\label{portfolio-opt}
\begin{array}{rr@{}ll}
\underset{w_f,x,z_1,z_2,z_3,f,y_\text{tickets},y_\text{names}}{\text{maximize}}  & \sum_{i=1}^n \alpha_i w(i) &\\

\text{subject to}
& \sum_{i=1}^n w_f(i) & = 1\\
& \sum_{s\in S}x(s) &\leq \sectormax \\
& \sum_{i=1}^n \ynames(i) &\leq \namemax \\
& \sum_{i=1}^n y_{\text{tickets}}(i) &\leq \ticketmax \\

& w_f(i) - w_t(i) & \leq y(i) & \irange\\
& -(w_f(i) - w_t(i)) & \leq y(i) & \irange\\

& w_f(i) - w_0(i) & \leq f(i) & \irange\\
& -(w_f(i) - w_0(i)) & \leq f(i) & \irange\\

& \sum_{i=1}^n M_s(i)(w_f(i) - w_t(i)) & \leq x(s) & \forall s\in S\\
& -\sum_{i=1}^n M_s(i)(w_f(i) - w_t(i)) & \leq x(s) & \forall s\in S\\

& w_f(i) & \leq \ynames(i) & \irange \\
& f(i) & \leq y_{\text{tickets}}(i) & \irange\\ 
& f(i) & = z_1(i) + z_2(i) + z_3(i) & \irange\\ 
& 0 \leq z_1(i) &\leq 0.1 \volume(i) & \irange\\
& 0 \leq z_2(i) &\leq 0.2 \volume(i) & \irange\\
& 0 \leq z_3(i) &\leq 0.2 \volume(i) & \irange\\
& w_f(i), f(i), y(i) & \geq 0 & \irange\\
& z_1(i), z_2(i), z_3(i) & \geq 0 & \irange\\
& \ynames(i), y_{\text{tickets}}(i) & \in \{0,1\} & \forall i =  1,\dots,n\\
& x(s) & \geq 0 & \forall s \in S
\end{array}
\end{equation}

Overall this problem has $\lvert S \rvert + 6n$ continuous decision variables, $\lvert 2n\rvert$ binary decision variables, and $10n+\lvert 2S \rvert + 4$ constraints.

\section{Bipartite matching}

\subsection{Data specification}
We run experiments on a version of the bipartite matching problem used in \cite{wilder2018melding} which requires additional diversity constraints on the proposed matching. The matching problem is done on graphs sampled from the CORA citation network dataset \cite{sen2008cora} which consist of nodes and edges representing publications and citations respectively. In addition, we use information present about a given paper's field of study. The full network we consider consists of 2708 nodes which are partitioned into 27 matching instances using metis \cite{karypis1998metis}. Each instance is a bipartite matching problem on a complete bipartite graph with 50 nodes on each side of the graph. The 100 nodes in each instance are divided to maximize the number of edges between the two partitions. To ensure the diversity of our recommendations, we require a minimum $p=25\%$ percentage of the suggested pairings belong to distinct fields of study and similarly impose a constraint that a minimum percentage $q=25\%$ of suggested citations belong to the same field of study. The predicted values in this case are the edge weights which correspond to whether one paper actually cites another. The node representations in this case are binary feature vectors which correspond to whether a given word from a lexicon of 1433 words appears in the paper. Edge values are predicted based on the concatenation of the node features on the edge endpoints. 

\subsection{Optimization model}
\textbf{Given:} two sets of nodes representing publications $N_1, N_2$, weights on pairs of nodes $c_{i,j} \forall i \in N_1, j\in N_2$ corresponding to how likely it is that one paper will cite another, same field indicator $m_{i,j}=\{1 \text{ if i and j are in the same field}, 0 \text{ otherwise}\}$\\
\textbf{Goal:} find matching of nodes such that 1) each node is matched at most once, 2) at least $p \in [0,1]$ proportion of selected edges connect nodes of the same field ($m_{i,j}=1$) 3) at least $q\in [0,1]$ proportion of selected edges connect nodes of different fields ($m_{i,j}=0$)\\
\textbf{Decision Variables:} 
$x_{i,j} \in{0,1} \forall (i,j)\in N_1 \times N_2$ corresponding to whether we use edge $(i,j)$ in the matching or not. \\
\textbf{Formulation:}
\begin{equation}
\label{matching-opt}
\begin{array}{lr@{}ll}
\underset{x}{\text{maximize}}  & \sum_{i,j}c_{i,j}x_{i,j} &\\
\text{subject to}
& \sum_{j}x_{i,j} & \leq 1 & \forall i \in N_1\\
& \sum_{i}x_{i,j} & \leq 1 & \forall j \in N_2\\
& \sum_{i,j} m_{i,j} x_{i,j} & \geq p \sum_{i,j} x_{i,j}\\
& \sum_{i,j} (1-m_{i,j}) x_{i,j} & \geq q \sum_{i,j} x_{i,j}\\
& x_{i,j} & \in {0,1} & \forall i\in N_1\,, j\in N_2
\end{array}
\end{equation}

Overall this problem has $\lvert N_1 \rvert \times \lvert N_2 \rvert$ decision variables, and $\lvert N_1 \rvert + \lvert N_2 \rvert + 2$ constraints. In our setting, we had $p=q=0.25$ to validate our experiments as this setting resulted in the problem not simply being solved too easily while yielding feasible regions.

\end{document}